\documentclass[letterpaper, 10 pt, conference]{ieeeconf}  
\IEEEoverridecommandlockouts                              

\usepackage{graphics}    
\usepackage{times}       
\usepackage{amsmath}     
\usepackage{amssymb}     
\usepackage{graphicx}
\usepackage[separate-uncertainty=true]{siunitx}%
\usepackage[caption=false, font=footnotesize]{subfig}
\usepackage{tabularx}

\usepackage[hidelinks]{hyperref} 

\usepackage{flushend}

\def\secref#1{Sec.~\ref{#1}}

\def\tabref#1{Tab.~\ref{#1}}
\def\eqref#1{Eq.~(\ref{#1})}

\newcommand\etal{\emph{et al. }}

\title{\LARGE \bf Learning Depth Vision-Based Personalized Robot Navigation\\From Dynamic Demonstrations in Virtual Reality}

\author{Jorge de Heuvel \and Nathan Corral \and Benedikt Kreis \and Jacobus Conradi \and Anne Driemel \and Maren Bennewitz
  \thanks{J. de Heuvel, N. Corral, B. Kreis, and M. Bennewitz are with
    the Humanoid Robots Lab,  J. Conradi and A. Driemel are with the
    Group for Algorithms and Complexity, University of
    Bonn, Germany. M. Bennewitz and A. Driemel are additionally
    with the Lamarr Institute for Machine Learning and Artificial
    Intelligence, Germany. This work has partially been funded by the Deutsche Forschungsgemeinschaft (DFG, German Research Foundation) under the grant number \mbox{BE~4420/2-2}~(FOR 2535 Anticipating Human Behavior).}
}

\begin{document}
\maketitle
\thispagestyle{empty} 
\pagestyle{empty}

\begin{abstract} 

For the best human-robot interaction experience, the robot's navigation policy should take into account personal preferences of the user.
In this paper, we present a learning framework complemented by a perception pipeline to train a depth vision-based, personalized navigation controller from user demonstrations.
Our virtual reality interface enables the demonstration of robot navigation trajectories under motion of the user for dynamic interaction scenarios.
The novel perception pipeline enrolls a variational autoencoder in combination with a motion predictor.
It compresses the perceived depth images to a latent state representation to enable efficient reasoning of the learning agent about the robot's dynamic environment.
In a detailed analysis and ablation study, we evaluate different configurations of the perception pipeline.
To further quantify the navigation controller's quality of personalization, we develop and apply a novel metric to measure preference reflection based on the Fréchet Distance.
We discuss the robot's navigation performance in various virtual scenes and demonstrate the first personalized robot navigation controller that solely relies on depth images.
A supplemental video highlighting our approach is available online\footnote{Full video: \href{https://www.hrl.uni-bonn.de/publications/deheuvel23iros_learning.mp4}{hrl.uni-bonn.de/publications/deheuvel23iros\_learning.mp4}}.

\end{abstract}

\section{Introduction}
\label{sec:intro}
The personalization of robots will be a key factor for comfortable and satisfying human-robot-interactions.
As the integration of robots at home or at work will inevitably increase, the number one goal should be a naturally collaborative experience between users and the robot.
However, users might have personal preferences about specific aspects of the robot's behavior that define the personal golden standard of interaction.
Falling short of user's preferences could lead to negative interaction experiences and consequently frustration~\cite{kruse_human-aware_2013}.

Where humans share the same environment with a mobile robot, the robot's navigation behavior significantly influences the comfort of interaction \cite{de_heuvel_learning_2022, francis_principles_2023}.
Consequently, basic obstacle avoidance approaches are insufficient to address individual preferences regarding proxemics, trajectory shape, or area of navigation in a given environment, while being a key component to successful navigation without question.
Instead, a robot's navigation policy should be aware of humans \cite{moller_survey_2021} and reflect the users' personal preferences.

In our previous work \cite{de_heuvel_learning_2022} we demonstrated that pairing a virtual reality (VR) interface with a reinforcement learning (RL) framework enables the demonstration and training of highly customizable navigation behaviors.
The resulting navigation controller outperformed non-personalized controllers in terms of perceived comfort and interaction experience.
However, a key assumption in the previous work is an always-present, static human of known pose in a predefined environment with pose-encoded obstacles.
This benefits the learning process with a low-dimensional state space.
To overcome these assumptions, enrolling a depth vision sensor to sense both human and obstacles is a possible solution~\cite{theodoridou_robot_2022}.
However, depth vision cameras come at the cost of high-dimensional, complex, and redundant output.
Learning from such high-dimensional data on dynamic scenes is a challenging task~\cite{laskin_curl_2020}.
The question crystallizes, how do we teach preferences of moving users in realistic environments, while relying on state-of-the art sensor modalities? 

To solve the challenges above, we introduce a depth vision-based perception pipeline that is both lightweight, human-aware and, most importantly, provides the robot with a low-dimensional representation of the dynamic scene.
This pipeline i) detects the human and obstacles, ii) compresses the perceived depth information, and iii) enables efficient reasoning about the robot's dynamic environment to the learning framework.
Our new system is able to learn personalized navigation preferences from a VR interface and learning framework for dynamic scenes in which both robot and human move.
\begin{figure}[t]
	\centering
	\includegraphics[width=0.90\linewidth]{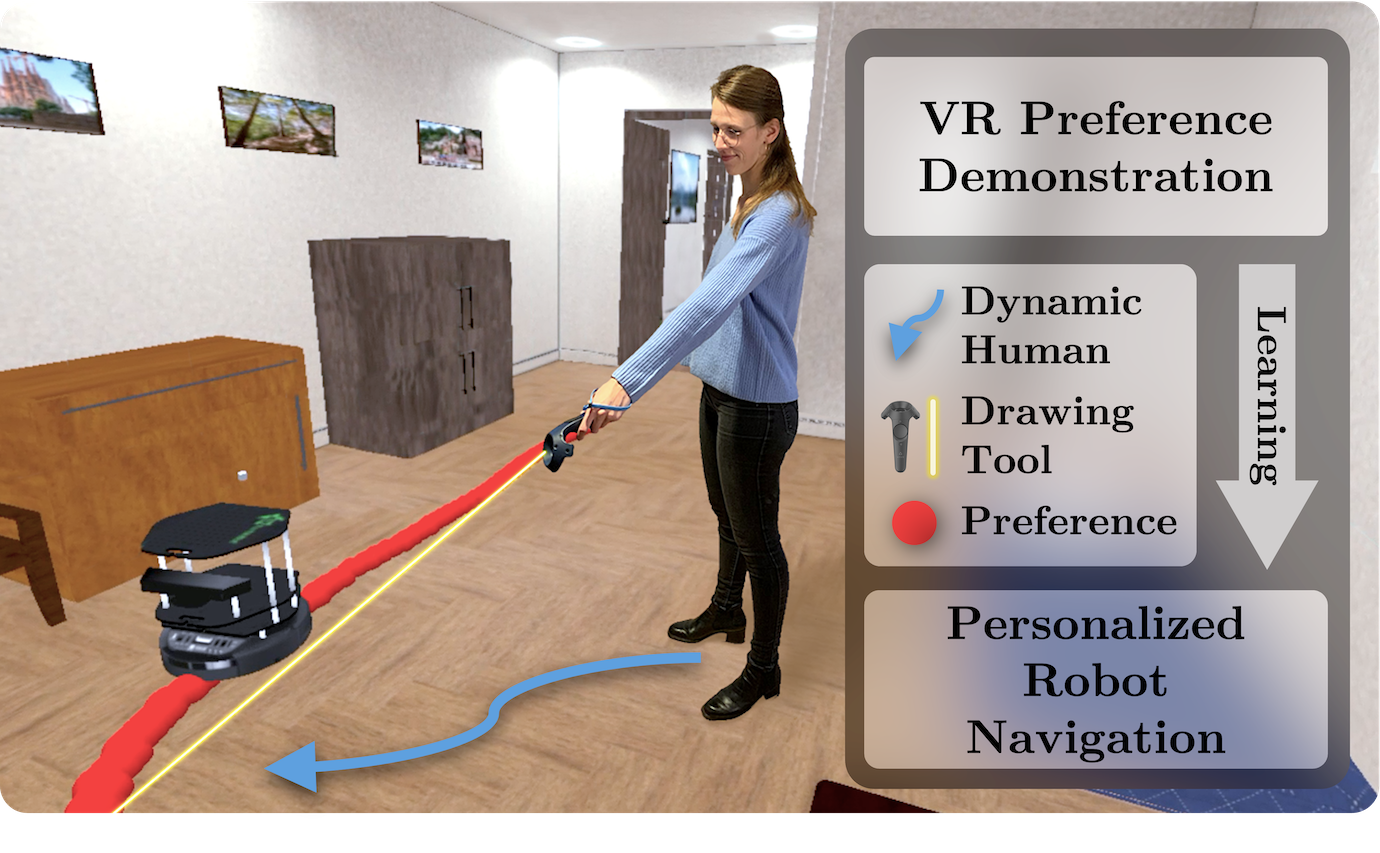}
	\caption{
		Our virtual reality~(VR) interface allows the demonstration of robot navigation preferences by drawing trajectories intuitively onto the floor. 
		By applying a learning-based framework, we achieve personalized navigation using a depth vision-based perception pipeline.
		\label{fig:motivational_add_on}}
\end{figure}
\begin{figure*}[ht!] 	\centering 	\includegraphics[width=0.96\linewidth]{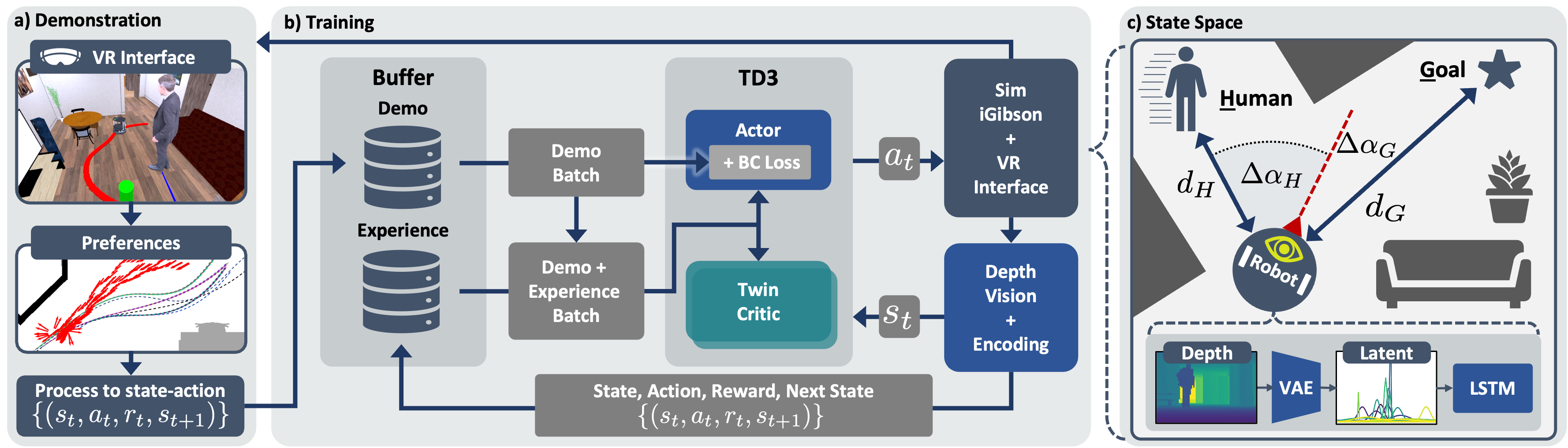} 	\caption{ 		Schematic representation of our architecture.
		\textbf{a)} Demonstration trajectories are drawn by the user in VR onto the floor using the handheld controller.
		Subsequently, the trajectories are fed into the demonstration buffer.
		\textbf{b)} Our TD3 reinforcement learning architecture with an additional behavioral cloning (BC) loss on the actor trains a personalized navigation policy that outputs linear and angular velocities.
		\textbf{c)} The robot-centric state space relies on a depth vision perception pipeline, capturing the vicinity of the human and obstacles in the environment, as well as the relative goal position.
		A variational autoencoder (VAE) compresses the raw images to a latent state representation, while a predictor (LSTM) provides subsequent state predictions.
	} 
	\label{fig:architecture}
\end{figure*} 

In summary, the \textbf{main contributions} of our work are: \begin{itemize} 
	\item Learning a preference-reflecting navigation controller that relies solely on depth vision.
	\item A VR demonstration framework to record navigation preferences for a dynamic human-robot scenario.
	\item The introduction and application of a novel metric to quantify the quality of navigation preference reflection.
	\item An extensive qualitative and quantitative analysis of different perception configurations for personalized navigation.
\end{itemize}

\section{Related Work}
\label{sec:related}
Adjusting or learning the navigation behavior of a robot based on feedback or demonstration has been the focus of various studies \cite{kollmitz_learning_2020, gao_modeling_2019, marta_aligning_2023}.
Especially, deep learning-based approaches shine by their ability to learn from subtle and implicit features in their environment \cite{pfeiffer_reinforced_2018, karnan_socially_2022, chen_crowd-robot_2019}.
This is an ideal motivation to use a deep RL architecture for our personalized navigation controller.

Fusing the potential of user demonstrations with a learning architecture led to promising results in the field of robotic manipulation tasks \cite{nair_overcoming_2018}.
Therefore, this is a key concept for our learning architecture and has successfully been applied to the field of robot navigation \cite{de_heuvel_learning_2022}.

Vision-based sensor modalities for navigation appeal due to their cost-efficiency.
For human-aware navigation, the detection and explicit localization of pedestrians enabled socially conforming navigation controllers \cite{theodoridou_robot_2022, tai_socially_2018}.

Recent advances in the field of depth vision-based navigation in combination with RL have been made by Hoeller~\etal \cite{hoeller_learning_2021}, who study a state representation of depth-images to efficiently learn navigation in dynamic environments.
Our proposed perception pipeline is built upon their successful architecture.

Furthermore, a navigating agent benefits from dynamic scene understanding.
Predicting the movement of surrounding pedestrians and obstacles with Long Short-Term Memory~(LSTM) models has lead to promising results \cite{alahi_social_2016, fernando_soft_2018, hoeller_learning_2021}.
Therefore, we will integrate an LSTM architecture into our perception pipeline.

While in our previous work~\cite{de_heuvel_learning_2022} we presented one of the first approaches at the intersection of navigation and robot personalization, we now enhance the system by allowing the user to demonstrate navigation trajectories under dynamic motions and using only depth vision as controller input.

\section{Our Approach}
\begin{figure*}[ht!] 	\centering 	\includegraphics[width=1.0\linewidth]{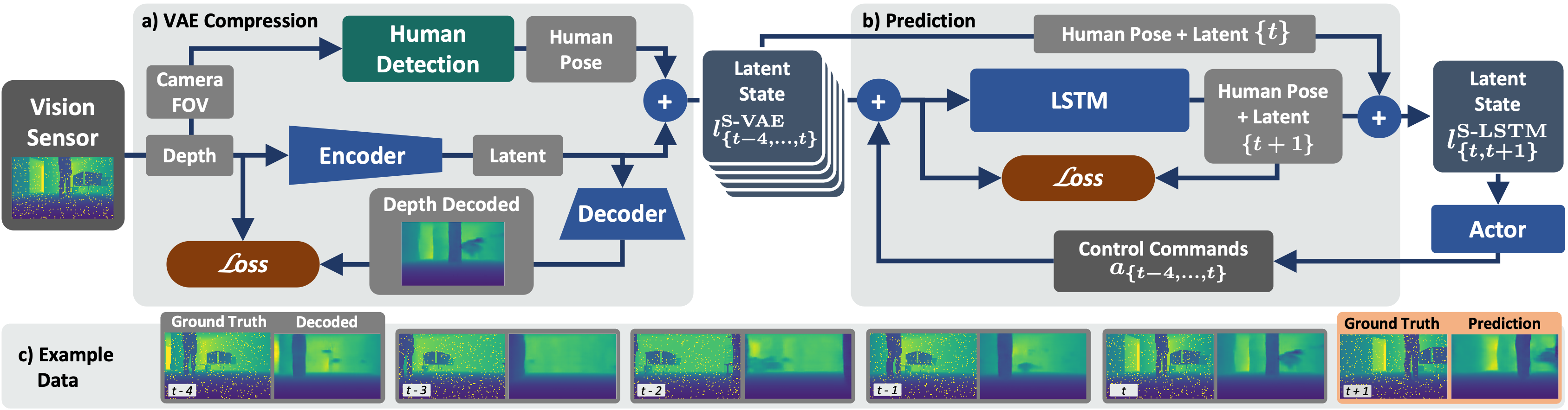} 	\caption{ 		Schematic representation of the perception pipeline.
		\textbf{a)} The vision sensor's depth frames are encoded using a VAE to a latent space of dimensionality~32.
		In parallel, we check for human presence in the sensor's field of view (FOV), in which case we provide the human position relative to the robot.
		The merged latent state S-VAE provides our first state representation for learning.
		\textbf{b)} After the last five states $l^{\text{S-VAE}}_{\{t-4,\dots, t\}}$ are merged with the robot control commands $a_{\{t-4,\dots, t\}}$, the LSTM predicts both the next latent as well as the next human pose $l^{\text{S-LSTM}}_{\{t+1\}}$.
		Only the human pose prediction~$(d_{H}^{t+1}, \Delta\alpha_{H}^{t+1})$ is merged with the previous latent $l_t$ and human pose $(d_{H}^{t}, \Delta\alpha_{H}^{t})$ to the state S-LSTM.
		Both state versions S-VAE and S-LSTM are used separately for training.
		\mbox{\textbf{c)} Visualization} of the trained VAE and LSTM model with ground truth depth data before encoding (left in box) and the decoder's reconstruction (right in box).
		The LSTM predicts the next latent and human pose, where the latent reconstruction is shown (orange box).
		\label{fig:architecture_vision}
	} \end{figure*} 
In this work, we consider a robot navigating in the same room as a single, human user.
The user has personal preferences about the way the robot circumnavigates him/her while pursuing a local goal in the same room.
Such preferences could lie in the approaching behavior or the robot's trajectory.
We assume the robot to be provided a local goal from a global planner.
The local goal could be a door on the opposite side of the current room to be traversed, or a location of interest in the same room.
Using such sparse local goals several meters apart, we provide the controller with the spatial and temporal freedom to navigate towards the goal in a user-preferred personalized manner.
The human shares the navigation space with the robot, whether being dynamic by walking through the room, or resting static.
To achieve preference-aligned and collision-free navigation behavior, the robot relies only on a depth vision camera to sense the distance to the human as well as obstacles.
We formulate personalized navigation as a learning task, where the robot learns a personalized controller outputting linear and angular velocity from VR demonstrations of the user.

\subsection{Learning Architecture}
The learning approach presented in this section is a hybrid of reinforcement learning and behavior cloning.

RL refers to the optimization of environment interactions, leading from state $s_t \rightarrow s_{t+1}$ that obey a Markov Decision Process.
The interacting agent receives a reward $r_t = r(s_t, a_t)$ for taking an action $a_t = \pi_{\phi}(s_t)$ at time step $t$ with respect to a policy $\pi_{\phi}$.
The tuples $\left(s_t, a_t, r_t, s_{t+1}\right)$ are referred to as state-action pairs.
The optimization goal is to maximize the overall return $R = \sum^{T}_{i=t} \gamma^{(i-t)}r_t$ of the $\gamma$-discounted rewards, onward from time step $t$.

Fig.~\ref{fig:architecture} depicts a schematic overview of our approach.
We enroll an off-policy twin-delayed deep deterministic policy gradient (TD3) reinforcement learning architecture \cite{fujimoto_addressing_2018}.
In short, two critics networks learn to estimate the value of the state-action distribution, the actor networks learns a policy $\pi(s_t) = a_t$ ideally leading to the highest expected return $R$.
All three networks are standard multi-layer perceptrons (MLP) and share the same architecture.
For policy updates, batches of training data $b_E$ are sampled from the experience buffer.
TD3's continuous action space ensures smooth robot control, as the actor network outputs linear and angular velocities as control commands.

An additional modification to the standard TD3 is a behavioral cloning loss $\mathcal{L}_\text{BC} = \sum_{i=1}^{b_D} || \pi_{\phi}(s_i) - a_i ||^2$ on the actor network provided with demonstration data in batches $b_D$ \cite{nair_overcoming_2018} from a separate static buffer containing navigation preferences collected in VR, see Fig.~\ref{fig:architecture}a-b.
The extended and $\lambda_{\text{BC/RL}}$-balanced loss on the actor is \mbox{$\nabla_{\phi} J_\text{total} = \lambda_{\text{RL}} \nabla_{\phi} J - \lambda_{\text{BC}} \nabla_{\phi} \mathcal{L}_\text{BC}$} with the actor's original policy gradient $\nabla_{\phi} J$.

By continuously sampling data from both buffers and applying the BC loss throughout the training, a navigation policy is learned that exhibits demonstration-like behavior whenever the navigation scenario allows.
At the same time, the policy generalizes to unknown states not covered by the demonstration data.

\subsection{Representation Learning}
\label{sec:representation_learning}
This section provides implementation and training details on our perception pipeline depicted in Fig.~\ref{fig:architecture_vision}.

\subsubsection{Variational Autoencoder} Reinforcement learning on raw high-dimensional vision data is unfeasible.
Ideally, a dimensionality-reduced state representation is used~\cite{hoeller_learning_2021}.
Thus, we compress the depth data to a latent representation $l$ using a $\beta$-variational autoencoder (VAE) with six relu-activated convolutional layers, see Fig.~\ref{fig:architecture_vision}a.
The dimensionality reduction is factor 320 from a 128 x 80 pixel depth image to a latent space of dimensionality 32.
To make the model robust against sensor noise that a depth camera would exhibit, we apply a \num{5}\text{ }\si{\percent} dropout noise to the depth frames during VAE training.
The VAE learns to filter the noise, as the VAE's reconstruction loss is computed between the decoded and the noise-free depth-frame.
A visualization of the VAE's performance is depicted in Fig.~\ref{fig:architecture_vision}c.

\subsubsection{Predictor} Originating from single depth frames, the latent space alone fails to capture dynamic scene information such as motion or human movement.
To leverage dynamic scene information such as the human motion for the navigation controller, a predictor is introduced, see Fig.~\ref{fig:architecture_vision}b.
The predictor receives the last five human poses, control commands, and latent frames $(d_{H}^i, \Delta\alpha_{H}^i, a_i, l_i)_ {i \in \{t-4,\dots, t\}}$ as input.
We predict the next human pose $(d_{H}^{t+1}, \Delta\alpha_{H}^{t+1})$ and the latent of the next time-step $l_{t+1}$.
The model consists of two LSTM layers with 64 units each, followed two linearly activated MLPs which output both mean $\mu_{t+1}$ and variance $\sigma_{t+1}$ as in the VAE, from which the latent prediction $l_{t+1}$ is sampled.
The human pose prediction is performed by a two-layer MLP from the LSTM-layer's output.
A visualization of the predictor's performance is depicted in Fig.~\ref{fig:architecture_vision}c.

\subsubsection{Training Data} To train the autoencoder and predictor, we generated an extensive dataset of depth-frames in the iGibson simulator~\cite{shen_igibson_2021}.
Here, we used the scene setup described in Sec.~\ref{sec:training} with a static or dynamic human.
The robot's navigation policy for the dataset generation was a simple obstacle avoidance controller trained with TD3 RL.
Furthermore, the dataset contains ground-truth data about the human pose and the human's presence in the RGB-D cameras's field of view~(FOV).

\subsection{State and Action Space}
\label{sec:statespace}
Our robot-centric state space consists of three main parts, compare Fig.~\ref{fig:architecture}c: 1) The relative goal position $(d_G, \Delta\alpha_G)$, 2) the human position $(d_H, \Delta\alpha_H)$ and presence $k_H \in \{0,1\}$ in the robot's FOV, and 3) the latent representation of the depth data.
The human state corresponds to the current time step $t$ for VAE-only configurations, and to a $t+1$ prediction concatenated with the $t$-human state for the VAE+LSTM.
Thus, the human is both implicitly encoded as an obstacle in the latent-encoded depth image, but also explicitly.
All positions are given in robot-centric polar coordinates.
When no human is observed in the FOV, then $d_{H}^*=-1$ m and \mbox{$\Delta\alpha_{H}^* = 0$~rad}.
The actor's action space is composed of forward and angular velocity $(v, \omega)$, which are used as control commands.

\subsection{Reward}
\label{sec:reward}
We aim to teach user-specific navigation preferences not by complex reward shaping, but only via demonstration data.
Consequently, we keep the reward as sparse as possible besides basic collision penalties and goal rewards 
\begin{align}
	r = r_\text{collision} + r_\text{goal} + r_\text{timeout} \text{.}
\end{align}

The scaling factor $c_\text{rew} = 10$ is used throughout the reward definition below.
Upon collision with either an obstacle or human, we penalize with \mbox{$r_\text{collision} = - \frac{1}{2} c_\text{rew}$}.
When the robot reaches the goal location, a positive reward is provided: 
\begin{align}
	\label{eq:reward_goal}
	r_\text{goal} = 
	\begin{cases} 
		+ c_\text{rew} & \text{if goal reached in demonstration data} \\
		+ \frac{c_\text{rew}}{2} & \text{if goal reached during training} \\
		0 & \text{else}
	\end{cases}
\end{align}

Note the explicitly higher reward of the demonstration data to boost the value of demonstration-like behavior for the critics during learning.
This is complemented further by an additional $+\frac{c_\text{rew}}{100}$ on each demonstration state reward.
In short, a higher value of demonstration-like behavior encourages user-preference-like navigation whenever possible, while preventing the agent from taking more efficient, shorter trajectories to achieve the faster and higher return $R$.

To overcome navigation behavior that does not lead to the goal on the long run, upon timeout when \mbox{$n > N_\text{ep}$} we penalize with $r_\text{timeout} = - \frac{c_\text{rew}}{4}$.
In all other cases the reward is zero.
An episode denotes the trajectory roll-out from initial robot placement until one of the termination criteria is satisfied.
All three reward criteria are also episode termination criteria.

\begin{figure*}[t] 	\centering 	\includegraphics[width=0.9\linewidth]{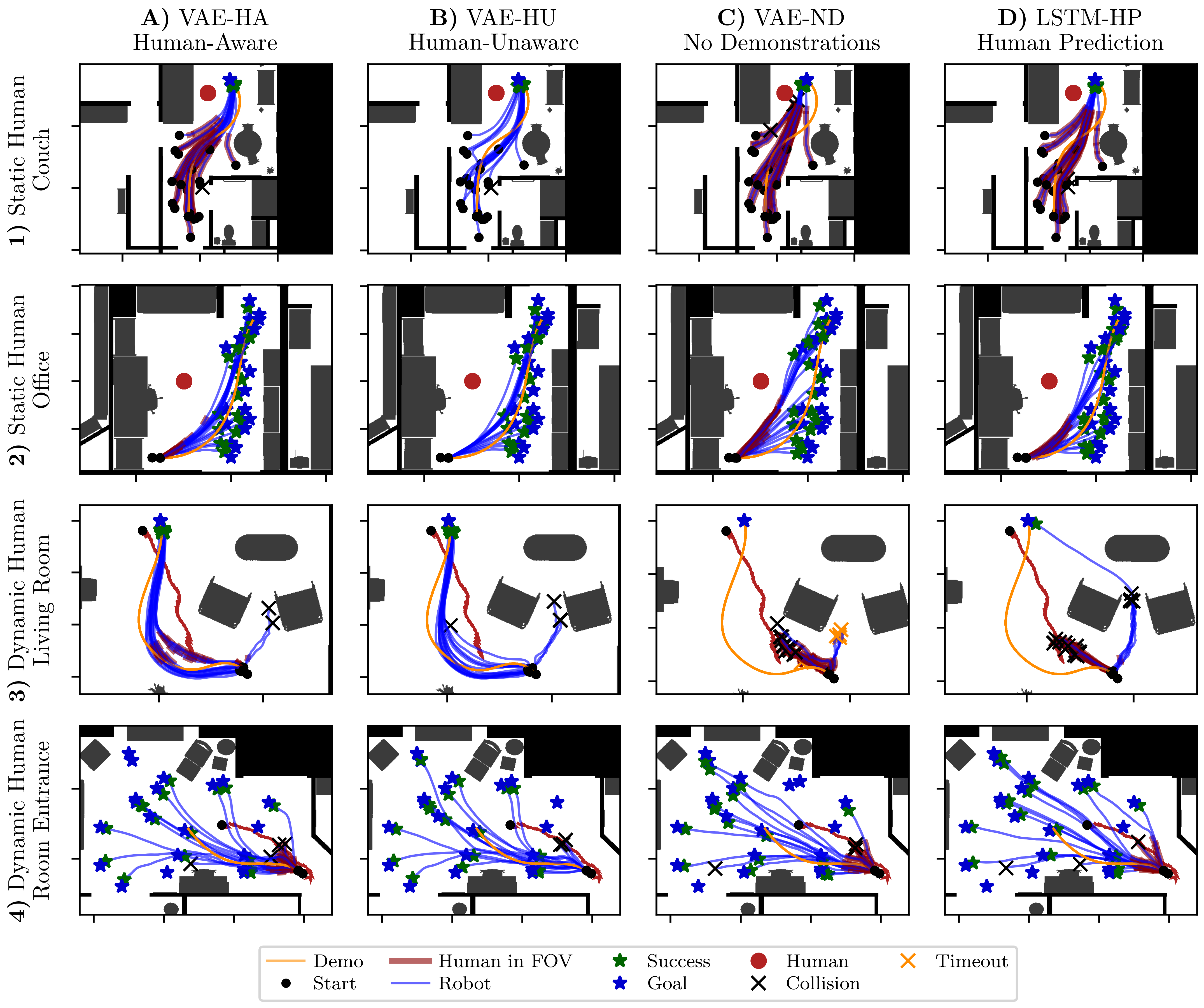} 	\caption{ 		The robot's learned navigation behavior (blue lines) for scenes, where preferences were demonstrated (\textbf{rows 1-4}) and perception as well as learning configurations (\textbf{columns A-D}) are depicted.
		For all scenes, one demonstrated navigation preference is shown (orange).
		The human (red) is either static (red circle) or moving through the scene (red arrow).
		Goal (blue star) and start location (black dot) are either taken from the demonstration trajectory or sampled equivalently in the room across all configurations.
		Whenever the human is in the RGB-D camera's FOV, the robot trajectory is shaded in red.
		In short, the VAE-HA approach (A) exhibits navigation behavior which is the closest to the demonstrated preferences.
		As the human-detection is turned off on the VAE-HU (B) and no human pose is provided to the controller, the robot performs less pronounced avoidance (B1 vs.
A1).
		In contrast, the VAE-ND controller trained without any demonstration data (C) rather reflects a shortest path driving behavior.
		In the most challenging scene (3) LSTM-HP (D) and VAE-ND (C) fail, where approach VAE-HA (A3) shines.
	} 
	\label{fig:analysis}
\end{figure*} 
\section{Demonstration and Training Environment}
We first introduce the advances on the VR interface, in which a human user teaches personal navigation preferences to a robot - now under dynamic motion.
Subsequently, the learning environment and navigation task are presented.

\vspace{-0.5em}
\subsection{Simulator and Robot}
To teach and train our navigation controller in a more realistic environment with RL, we use the iGibson simulator~\cite{shen_igibson_2021} that provides a set of interactive indoor scenes and a VR interface that we used for immersive demonstration.
iGibson renders the robot's vision sensors, which serve as input to our perception pipeline during training.
Its underlying physics engine is Pybullet \cite{coumans_pybullet_2016}.
We focus on the differential-wheeled robot Kobuki Turtlebot 2.
Generally, our approach is applicable to other robots with similar control modalities.
The Turtlebot's control limits lie at $v \in [\num{0},\num{0.5}] \text{ }\si{\meter\per\second}$ forward and $\omega \in [-\pi, +\pi] \text{ }\si{\radian\per\second}$ angular.
Inspired by the Intel Realsense D455 depth camera, the robot features a forward facing depth-camera with a \num{87}\si{\degree} horizontal FOV.
The depth-sensing range is limited to $\num{6}\text{ }\si{\meter}$, which is equivalent to a temporal foresight of $\num{12}\text{ }\si{\second}$ at the Turtlebot's maximum forward velocity.
As there is no sensor facing backward to sense rear obstacles, the Turtlebot is not allowed to drive backward.

\subsection{Collecting and Processing Demonstration Trajectories}
\label{sec:demonstration}
In the VR preference collection procedure, the demonstrating user can now move and teach dynamic situations, as compared to \cite{de_heuvel_learning_2022}.
To demonstrate, the user firstly familiarizes him/herself with the environment in VR.
Subsequently, he/she demonstrates a trajectory for the robot by drawing it onto the floor using the beam-emitting handheld controller, see Fig.~\ref{fig:motivational_add_on}.
There is no preset goal for the robot in the demonstration scene, so the user can demonstrate preferences in any direction.
The goal location will automatically be set to the end of the demonstration trajectory.
As the robot executes the trajectory from analytically computed action commands, the evolution of the human position and orientation is recorded from the wireless head-mounted display's location.
So to complement the demonstration with his/her movement, the user can walk freely in the scene while the robot navigates.
Just like the robot's trajectory, also the human trajectory is converted into a spline representation to be replayed during training and when the state-action pairs for the demonstration buffer are subsequently recorded.
In a last step, the user can step aside and observe the moving robot and human 3D~mesh from a third-person perspective.
The demonstrations are double-checked for any collisions that would result in negative rewards upon replay to the demonstration buffer to ensure their quality for the learning process.

The conducted user study in our previous work \cite{de_heuvel_learning_2022} has shown great acceptance of the VR interface and perceived navigation comfort of the learned controller.
In this work, we focus on the development and evaluation of a depth vision-based perception pipeline.
For this study, we recorded dynamic and static navigation scenarios by ourselves.
The dataset contains nine scene configurations, with around three demonstration trajectories each.

\subsection{Navigation Task and Training}
\label{sec:training}
We train our navigation controller on a set of interactive iGibson scenes and demonstration scenarios.
Start and goal location of the robot are randomly sampled in the same room, while ensuring a goal distance $d_G$ between $\num{1.5}\text{ }\si{\meter} < d < \num{6}\text{ }\si{\meter}$, equivalent to the depth sensing range.

To simulate the human in the scene, four different behaviors modes are sampled: 1) Human walks in the opposite direction from the robot's goal to its start on an A* path, thus encountering the robot.
2) Random human start and goal location.
3) The human is static.
4) No human in scene.
5) Human moves according to recorded demonstrations.
For modes 1+2, the human speed is sampled from a standard distribution $\mathcal{N}(\mu = 0.5\text{ }\si{\meter\per\second}, \sigma= 0.3\text{ }\si{\meter\per\second})$.

Lastly, we randomize over a set of iGibson scenes during training and change scenes every 50 episodes.

Before training begins, the experience buffer is initialized with $\num{5e+3}$ samples by executing randomly sampled actions.
An overview on all relevant and experimentally obtained training parameters can be found in \tabref{tab:training_settings}.

\section{Experimental Evaluation}
\label{sec:exp}
This section highlights the performance of our learned preference-reflecting navigation controller under different configurations.
A qualitative analysis in Sec.~\ref{sec:qualitative_analyis} showcases and discusses the navigation behavior on a set of selected scenes.
This is followed by a quantitative analysis targeting the robustness with success metrics in Sec.~\ref{sec:quantitative_analyis}.
Lastly in Sec.~\ref{sec:frechet}, we introduce a customized Fréchet similarity metric to quantify the quality of preference reflecting navigation behavior with respect to the demonstrations.

\vspace{+1.5em}
\begin{table}[!b]
	\centering
	\caption{Notations and Training settings. \label{tab:training_settings}}
	\begin{tabularx}{\linewidth}{llX}
		Notation & Value & Description \\
		\hline 
		$\beta$ & 3 & Weighting factor of the VAE's KL-divergence\\
		$N_\text{ep}$ & 150 & Maximum number of steps per episode\\
		$B_{E}$ & \num{2e+5} & Experience replay buffer size\\
		$b_{E/D}$ & 64 & Batch size of experience/demo data\\
		$l_a$ & \num{1e-4}  & Learning rate of actor\\
		$l_c$ & \num{8e-4}  & Learning rate of critic\\
		$\gamma$ & 0.99 & Discount factor\\
		$\sigma_{\epsilon_\pi}$ & 0.2 & Std. deviation of exploration noise $\epsilon_\pi$ \\
		$\sigma_{\epsilon_\theta}$ & 0.05 & Std. deviation of target policy noise $\epsilon_\theta$ \\
		$\lambda_{\text{RL}}$ & 30/4 & Weighting factor of RL gradient on actor\\
		$\lambda_{\text{BC}}$ & 10/4 & Weighting factor of BC loss gradient on actor\\		
		\hline
	\end{tabularx}
\end{table}
\subsection{Perception Pipeline Configurations}
We first evaluate different perception pipeline and learning configurations against each other, compare Fig.~\ref{fig:analysis}.A-D and Fig.~\ref{fig:ablation_study}.A-C.
Their key differences lie in the state space as input to the RL policy.

The standard \textbf{h}uman-\textbf{a}ware VAE-HA (Fig.~\ref{fig:analysis}A) state space configuration S-VAE contains the current latent depth encoding, goal position, the human presence binary and human position: $s_t^\text{VAE-HA} = (l_t, d_G, \Delta\alpha_G, k_H^t, d_H^t, \Delta\alpha_H^t)$.

The \textbf{h}uman-\textbf{u}naware VAE-HU (Fig.~\ref{fig:analysis}B) is the same controller as the VAE-HA, but the human detection in the robot's field of view is disabled during evaluation.

The \textbf{n}o-\textbf{d}emonstration VAE-ND controller does not rely on the learning architecture as shown in Fig.~\ref{fig:architecture}.
It has neither a demonstration buffer, nor a behavioral cloning loss, making it a standard TD3 architecture.
Therefore, it has learned its navigation behavior without user demonstrations.

The \textbf{h}uman-\textbf{p}rediction LSTM-HP (Fig.~\ref{fig:analysis}D) state space configuration S-LSTM is similar to S-VAE, except for the additional prediction of the next human position: $s_t^\text{LSTM-HP} = (s_t^\text{VAE-HA}, d_H^{t+1}, \Delta\alpha_H^{t+1})$.
Therefore, S-LSTM provides a dynamic scene information by predicting the human movement.

Our ablation study introduces two more configurations, see Sec.~\ref{sec:ablation}: VAE-FOV-120 implements a widened FOV at $120^\circ$ over the standard $87^\circ$, as it can be found on wide-angle depth cameras such as the Microsoft Azure Kinect.
VAE-NG discards the goal distance $d_G$ from the state space.

\subsection{Qualitative Navigation Analysis}
\label{sec:qualitative_analyis}
Fig.~\ref{fig:analysis} shows the learned navigation behavior of our controller and highlights resulting differences between the perception pipeline configurations introduced above.

In \textbf{Fig.~\ref{fig:analysis}.1}, the human is static and located at the couch.
The robot's start location is randomized, while keeping the goal at the end of the demonstration trajectory.
As the robot traverses the living room, it shall navigate on the opposite side of the room close to the dining table and along the cupboard.
With VAE-HA, the robot learned to navigate closely to the demonstrated preference.
It exhibits a similar, smooth, S-shaped curve while passing by the couch.
Interestingly, only little difference in the robot's overall trajectory shape can be observed between VAE-HA and VAE-HU (Fig.~\ref{fig:analysis}.A1+B1).
Here, a few trajectories traverse closer to the human (red dot).
So even though the human is not explicitly observed in the state space of VAE-HU, its overall approaching behavior to the human still reflects demonstration patterns.
A possible explanation is the agent's anchoring of behavior to the overall scene layout, rather than the human position.
Note that the robot trajectories are shaded in red in Fig.~\ref{fig:analysis}, whenever the human is observed on tho FOV.

Located at the desk in \textbf{Fig.~\ref{fig:analysis}.2}, the static human prefers the robot to take a wide turn as it leaves the corner next to the desk.
In this scenario, the navigation behavior among all configurations except VAE-ND is mostly reflecting the wide turn, where VAE-ND cuts short on the wide turn as expected.

In \textbf{Fig.~\ref{fig:analysis}.3}, the moving human encounters the robot with an opposite direction of travel at the living room's suite.
As a preference, the robot should take a wide turn of avoidance around the armchair to make space for the approaching human.
Among all controllers but VAE-HA, the navigation of the situation is challenging, leading to collisions around the armchair's corner.
While LSTM-HP fails to exhibit the demonstrated behavior in this scenario, VAE-HA and -HU display successful preference-like behavior in most cases.

As the human walks out of the room in \textbf{Fig.~\ref{fig:analysis}.4}, the robot enters.
Upon detection of the approaching human, the robot shall take a left turn and make room for the human to pass.
Afterwards, the robot can continue traversing the living room to its goal.
In this scenario the effect of demonstration trajectories strikes: The VAE-ND controller without access to demonstrations mostly exhibits direct goal-oriented, straight-path navigation.
Interestingly, the same applies for the LSTM-HP controller, while it exhibits superior collision avoidance towards the approaching human over VAE-HA, though with some reduction in preference reflection.

Qualitatively speaking, the VAE-HA configuration results in the best-performing personalized robot navigation controller.
Interestingly, the LSTM-HP configuration does not seem to provide a significant improvement compared to VAE-HA in most cases, but presumably at the cost of weaker preference reflection, compare Sec.~\ref{sec:frechet}.

\subsection{Quantitative Analysis: Robustness}
\label{sec:quantitative_analyis}
\begin{figure}[t]
	\centering
	\includegraphics[width=1.0\linewidth]{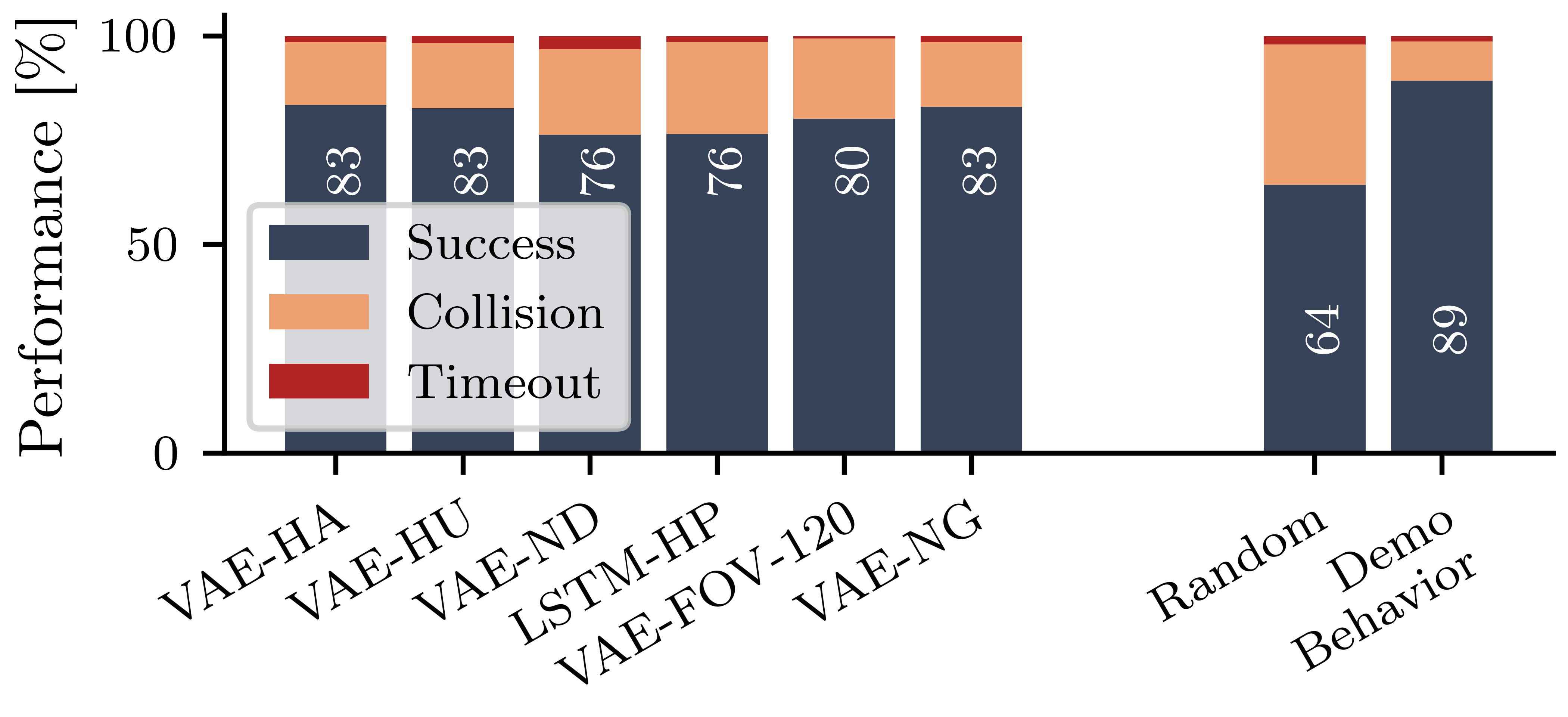}
	\caption{
		The performance of the different controllers is averaged over all demonstration scenarios and other scenes.
		For each combination of scene, human behavior modes, and demonstration preferences (if available), 50~trajectories were generated.
		"Random behavior" refers to behavior modes~1-4, while "demo behavior" refers to mode 5, both evaluated with controller VAE-HA.
		The success rates are shown on the plotted bars.
		\label{fig:performance}
	}
\end{figure}
Fig.~\ref{fig:performance} shows the performance of our different controller setups and human behavior modes (see Sec.~\ref{sec:training}) in terms of success rate, collision rate, and timeout rate.
We determine the demonstration-aware VAE architectures (VAE-HA, -HU, -NG) most capable of avoiding collisions with scene objects and the dynamic user.
Both the VAE-ND without demonstration access and the LSTM-HP controller perform worse than the demonstration-based VAE architectures.
Regarding different human behavior sampling modes~(\secref{sec:training}), as expected the demonstration-related mode 5 perform best.
We can also conclude from VAE-FOV-120 that increasing the RGB-D camera's field of view, e.g., for better perception of pedestrians approaching from the side, does not lead to better collision avoidance.
Generally, we observe more collision than timeout events.
This could be a consequence of the agent being encouraged to drive by the behavioral cloning loss from demonstration data.

\subsection{Quantitative Analysis: Preference Reflection}
\label{sec:frechet}
\begin{figure}[t]
	\centering
	\includegraphics[width=1.0\linewidth]{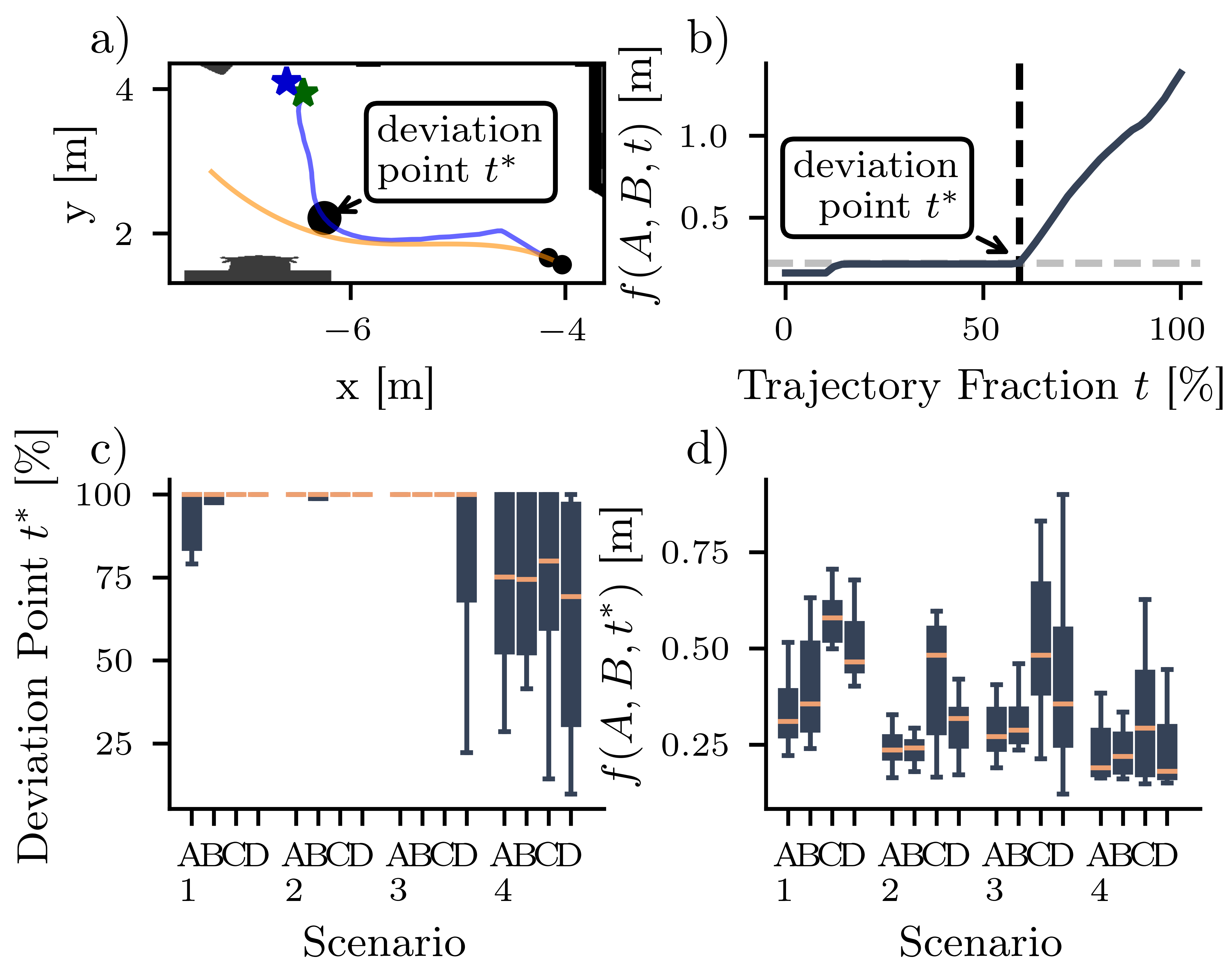}
	\caption{
		Visualization of our deviation-aware Fréchet metric $f(A, B, t^*)$.
		\textbf{a)} The robot follows the demonstrated path up to the deviation point $t^*$.
		Only up to this point we can reasonably compute a similarity between both trajectories.
		\textbf{b)} The deviation point $t^*$ is determined by the sudden increase in the Fréchet distance between demonstration and $t$-partially considered navigation trajectory via a cost function.
		\textbf{c)} With regards to all scenarios and trajectories in Fig.~\ref{fig:analysis}, the distribution of $t^*$ is shown.
		\textbf{d)} Consequently, the deviation-aware Fréchet metric $f(A, B, t^*)$ is computed, pointing towards the best and worst preference reflecting controller, \mbox{VAE-HA} (A) and \mbox{LSTM-HP} (D), respectively.
		\label{fig:frechet}
	}
\end{figure}
To quantify how closely the individual controllers reproduce the demonstrated preferences, we use the Fréchet distance between the navigation and demonstration trajectories.
The Fréchet distance $F(A, B)$ measures the similarity of two trajectories $A$ and $B$ \cite{alt_computing_1995}, by calculating the minimum value of the maximum distance between points on two curves or ordered points as \mbox{$F(A, B) = \inf_{\alpha, \beta} \max_{t\in[0, 1]} \left\| A(\alpha(t))-B(\beta(t))\right\|$}.
The order of points is taken into account with all possible reparameterizations $\alpha$ and $\beta$ of the curves, respectively.
We leverage the Fréchet distance not only to compute the similarity, but to estimate and quantify the point along the robot trajectory, where the robot significantly starts to deviate from the preference trajectory, as described below.
An example of this procedure is shown in Fig.~\ref{fig:frechet}a and \ref{fig:frechet}b on a given set of trajectories.
Firstly, the Fréchet distance is computed as a function of the considered fraction $t\in[0, 1]$ of the partial robot navigation trajectory $A[0, t]$ as 
\begin{align}
	\label{eq:frechet_evolution}
	f(A, B, t) = \inf_{t' \in [0, 1]} F(A[0,t], B[0, t'])\text{.}
\end{align}
Secondly, a trade-off cost $C_\varphi(t)$ between $f(A, B, t)$ and $t$ is computed as $C_\varphi(t) = \cos(\varphi) F(A, B, t) + \sin(\varphi) t$, where $\varphi = \frac{3}{4}\pi$.
Thirdly, we define the deviation point $t^*$ on trajectory $A$, where $\min_{t\in[0, 1]} C_\varphi(t)$.
In other words we estimate the point along the robot trajectory $t^*$, when $f(A, B, t)$ starts to continuously increase as the robot leaves the demonstrated path to pursue a goal aside the preference path.
Finally, we can determine, how closely the robot navigated along the demonstration trajectory up to the deviation point $t^*$, by evaluating $f(A, B, t^*)$.
We call $f(A, B, t^*)$ the deviation-aware Fréchet distance.
In Fig.~\ref{fig:frechet}a+b the deviation point is marked in both plots.

By applying our metric we solve the problem of either non-matching start or goal point between navigation and demonstration trajectory for a classical Fréchet analysis.
For those cases it would be pointless to quantify the similarity of both full-length trajectories with the plain Fréchet distance, as the deviation either at the end (same start) or at the beginning (same goal) would overshadow any measurable similarity.
Our deviation-aware Fréchet metric $f(A, B, t^*)$ calculates the Fréchet distance in an isolated manner on trajectory segments, among which similarity can be expected.
When the end-points are close instead of start points such as in Fig.~\ref{fig:analysis}.1, the metric is applied on the reversed trajectories $A$ and $B$.

We apply our deviation-point Fréchet metric to all navigation scenarios in Fig.~\ref{fig:analysis}.
On the one hand, we evaluated the deviation point $t^*$ in Fig.~\ref{fig:frechet}c and the corresponding deviation-aware Fréchet distance $f(A, B, t^*)$ in Fig.~\ref{fig:frechet}d.
We find the majority of navigation scenarios in Fig.~\ref{fig:analysis}.1-3 to fully follow the demonstration trajectory, which manifests in a deviation point $t^*$ close to $100\text{ }\si{\percent}$.
This is especially true for Fig.~\ref{fig:analysis}.3, where start and goal of navigation and demonstration overlap.
For the dynamic room entrance (Fig.~\ref{fig:analysis}.4), the robot's deviations from the demonstration path in favor for aside or further in the room positioned goals reflect in a lower distribution of $t^*$, see Fig.~\ref{fig:frechet}c.4A-D.
However, no obvious difference in $t^*$ can be observed when comparing the controller configurations, see Fig.~\ref{fig:frechet}c.A-D.
Interestingly, a clear difference in the deviation-aware Fréchet distance for the controller configurations can be found, see Fig.~\ref{fig:frechet}d.
Over all four navigation scenarios, controller VAE-HA (A in Fig.~\ref{fig:frechet}d and Fig.~\ref{fig:analysis}) exhibits the smallest deviation-aware Fréchet distance between preference and resulting navigation.
As expected, the worst preference-reflection can be assigned to the plain TD3 architecture without demonstration access VAE-ND (C in Fig.~\ref{fig:frechet}d and Fig.~\ref{fig:analysis}).

\subsection{Ablation Study}
\label{sec:ablation}
\begin{figure*}[t] 	\centering 	\includegraphics[width=0.93\linewidth]{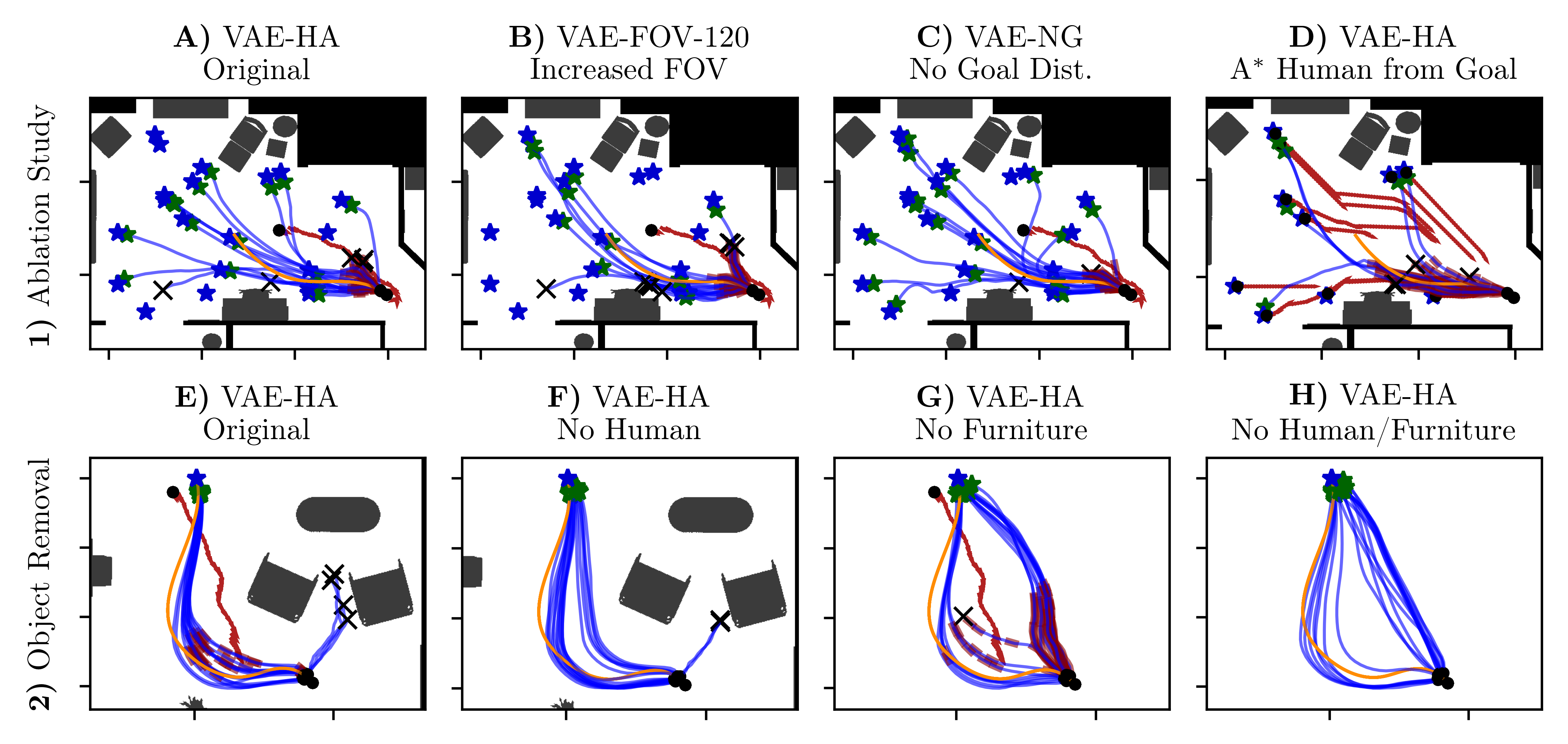} 	\caption{ 		\textbf{1)} In our ablation study, we investigate (\textbf{B}) the effect of increased camera field of view with VAE-FOV-120, (\textbf{C}) the removal of the goal distance from the state space VAE-NG in comparison to the original approach (\textbf{A}).
		\textbf{2)} To learn about relevant environment features for the agent, the human (\textbf{F}), the furniture (\textbf{G}), or both (\textbf{H}) were removed from the scene, compared to the original setup (\textbf{E}).
		For a legend, please refer to Fig.~\ref{fig:analysis}.
	} 
	\label{fig:ablation_study}
\end{figure*} 
Finally, we perform an ablation study to investigate effects of an increased camera field of view (Fig.~\ref{fig:ablation_study}.B) and the removal of the goal-distance from the state space (Fig.~\ref{fig:ablation_study}.C).
In the given scenarios, VAE-FOV-120 rather deteriorates the collision avoidance capabilities.
This is in line with the obtained overall performance results, see Fig.~\ref{fig:performance}.
Removing the goal distance (VAE-NG) interestingly does not deteriorate the performance, but also results in robust and preference-reflecting navigation.

Demonstrating the ability for generalization, in Fig.~\ref{fig:ablation_study}.D we showcase a scenario where humans follows an A* path in the opposite direction to the robot (compare behavior mode 1 in Sec.~\ref{sec:training}).
In most cases, the robot intuitively gives way to the approaching human.

To learn which features of the environment the agent uses for navigation and preference reproduction, we removed either the human, furniture, or both from the scene, see Fig.~\ref{fig:ablation_study}.E-H.
Interestingly, as no human approaches from behind the armchair (Fig.~\ref{fig:ablation_study}), the robot navigates closer to the chair with similar trajectory shape.
As all furniture is removed from the scene (Fig.~\ref{fig:ablation_study}.G), the robot either exhibits preference navigation or a shorter path on the other side of the approaching human.
With everything removed (Fig.~\ref{fig:ablation_study}.H), the small deviation around the human collapses to a shortest path on most trajectories.
Here, the deciding factor might be the initial orientation.
But also when neither human or furniture are part of the scene, the robot is able to reflect preferences.
We attribute this behavior to the walls and room layout that are still observable for the robot, or the learned guidance by relative goal position in the state space.

\section{Conclusion}
\label{sec:conclusion}
To summarize, we presented a learning approach to personalized navigation based on depth vision.
As demonstrated with our results, we successfully learned a personalized navigation controller that reflects user preferences from few VR demonstrations in dynamic human-robot navigation scenarios.
While various configurations have been tested, the extensive analysis points towards a pure VAE perception architecture for the best results.
Interestingly, including the motion predictor did not significantly improve the navigation performance or preference reflection.
Alongside the analysis, we have also developed and successfully applied a new metric that allows to quantify the quality of preference-reflection during navigation.
In conclusion, our research has demonstrated the feasibility of personalized robot navigation utilizing depth vision sensors and presents a promising avenue for further development.

\section*{Acknowledgments}
We gratefully thank Marlene Wessels and Camila Maslatón for their contribution on the motivational figure.

\bibliographystyle{IEEEtran}
\bibliography{selected_bib}

\begin{thebibliography}{10}
\providecommand{\url}[1]{#1}
\csname url@rmstyle\endcsname
\providecommand{\newblock}{\relax}
\providecommand{\bibinfo}[2]{#2}
\providecommand\BIBentrySTDinterwordspacing{\spaceskip=0pt\relax}
\providecommand\BIBentryALTinterwordstretchfactor{4}
\providecommand\BIBentryALTinterwordspacing{\spaceskip=\fontdimen2\font plus
\BIBentryALTinterwordstretchfactor\fontdimen3\font minus
  \fontdimen4\font\relax}
\providecommand\BIBforeignlanguage[2]{{%
\expandafter\ifx\csname l@#1\endcsname\relax
\typeout{** WARNING: IEEEtran.bst: No hyphenation pattern has been}%
\typeout{** loaded for the language `#1'. Using the pattern for}%
\typeout{** the default language instead.}%
\else
\language=\csname l@#1\endcsname
\fi
#2}}

\bibitem{kruse_human-aware_2013}
T.~Kruse, A.~K. Pandey, R.~Alami, and A.~Kirsch, ``Human-aware robot
  navigation: {{A}} survey,'' \emph{Robotics and Autonomous Systems}, vol.~61,
  no.~12, Dec. 2013.

\bibitem{de_heuvel_learning_2022}
J.~{de Heuvel}, N.~Corral, L.~Bruckschen, and M.~Bennewitz, ``Learning
  {{Personalized Human-Aware Robot Navigation Using Virtual Reality
  Demonstrations}} from a {{User Study}},'' in \emph{2022 31th {{IEEE Int}}.
  {{Conf}}. {{Robot Hum}}. {{Interact}}. {{Commun}}. {{RO-MAN}}}, 2022.

\bibitem{francis_principles_2023}
A.~Francis, C.~{Perez-D'Arpino}, C.~Li, F.~Xia, A.~Alahi, R.~Alami, A.~Bera,
  A.~Biswas, J.~Biswas, R.~Chandra, H.-T.~L. Chiang, M.~Everett, S.~Ha,
  J.~Hart, J.~P. How, H.~Karnan, T.-W.~E. Lee, L.~J. Manso, R.~Mirksy, S.~Pirk,
  P.~T. Singamaneni, P.~Stone, A.~V. Taylor, P.~Trautman, N.~Tsoi, M.~Vazquez,
  X.~Xiao, P.~Xu, N.~Yokoyama, A.~Toshev, and R.~{Martin-Martin}, ``Principles
  and {{Guidelines}} for {{Evaluating Social Robot Navigation Algorithms}},''
  \emph{arXiv:2306.16740 [cs]}, June 2023.

\bibitem{moller_survey_2021}
R.~M{\"o}ller, A.~Furnari, S.~Battiato, A.~H{\"a}rm{\"a}, and G.~M. Farinella,
  ``A survey on human-aware robot navigation,'' \emph{Robotics and Autonomous
  Systems}, vol. 145, Nov. 2021.

\bibitem{theodoridou_robot_2022}
C.~Theodoridou, D.~Antonopoulos, A.~Kargakos, I.~Kostavelis, D.~Giakoumis, and
  D.~Tzovaras, ``Robot {{Navigation}} in {{Human Populated Unknown
  Environments}} based on {{Visual-Laser Sensor Fusion}},'' in \emph{The15th
  {{Int}}. {{Conf}}. {{PErvasive Technol}}. {{Relat}}. {{Assist}}.
  {{Environ}}.}\hskip 1em plus 0.5em minus 0.4em\relax {ACM}, June 2022.

\bibitem{laskin_curl_2020}
M.~Laskin, A.~Srinivas, and P.~Abbeel, ``{{CURL}}: {{Contrastive Unsupervised
  Representations}} for {{Reinforcement Learning}},'' in \emph{Proc. 37th
  {{Int}}. {{Conf}}. {{Mach}}. {{Learn}}.}\hskip 1em plus 0.5em minus
  0.4em\relax {PMLR}, Nov. 2020.

\bibitem{kollmitz_learning_2020}
M.~Kollmitz, T.~Koller, J.~Boedecker, and W.~Burgard, ``Learning {{Human-Aware
  Robot Navigation}} from {{Physical Interaction}} via {{Inverse Reinforcement
  Learning}},'' in \emph{2020 {{IEEERSJ Int}}. {{Conf}}. {{Intell}}. {{Robots
  Syst}}. {{IROS}}}.\hskip 1em plus 0.5em minus 0.4em\relax {IEEE}, Oct. 2020.

\bibitem{gao_modeling_2019}
X.~Gao, X.~Zhao, and M.~Tan, ``Modeling {{Socially Normative Navigation
  Behaviors}} from {{Demonstrations}} with {{Inverse Reinforcement
  Learning}},'' in \emph{2019 {{IEEE}} 15th {{Int}}. {{Conf}}. {{Autom}}.
  {{Sci}}. {{Eng}}. {{CASE}}}.\hskip 1em plus 0.5em minus 0.4em\relax {IEEE},
  Aug. 2019.

\bibitem{marta_aligning_2023}
D.~Marta, S.~Holk, C.~Pek, J.~Tumova, and I.~Leite, ``Aligning {{Human
  Preferences}} with {{Baseline Objectives}} in {{Reinforcement Learning}},''
  in \emph{2023 {{IEEE Int}}. {{Conf}}. {{Robot}}. {{Autom}}. {{ICRA}}}, May
  2023.

\bibitem{pfeiffer_reinforced_2018}
M.~Pfeiffer, S.~Shukla, M.~Turchetta, C.~Cadena, A.~Krause, R.~Siegwart, and
  J.~Nieto, ``Reinforced {{Imitation}}: {{Sample Efficient Deep Reinforcement
  Learning}} for {{Mapless Navigation}} by {{Leveraging Prior
  Demonstrations}},'' \emph{IEEE Robot. Autom. Lett.}, vol.~3, no.~4, Oct.
  2018.

\bibitem{karnan_socially_2022}
H.~Karnan, A.~Nair, X.~Xiao, G.~Warnell, S.~Pirk, A.~Toshev, J.~Hart,
  J.~Biswas, and P.~Stone, ``Socially {{CompliAnt Navigation Dataset}}
  ({{SCAND}}): {{A Large-Scale Dataset}} of {{Demonstrations}} for {{Social
  Navigation}},'' \emph{IEEE Robot. Autom. Lett.}, vol.~7, no.~4, Oct. 2022.

\bibitem{chen_crowd-robot_2019}
C.~Chen, Y.~Liu, S.~Kreiss, and A.~Alahi, ``Crowd-{{Robot Interaction}}:
  {{Crowd-Aware Robot Navigation With Attention-Based Deep Reinforcement
  Learning}},'' in \emph{2019 {{Int}}. {{Conf}}. {{Robot}}. {{Autom}}.
  {{ICRA}}}, May 2019.

\bibitem{nair_overcoming_2018}
A.~Nair, B.~McGrew, M.~Andrychowicz, W.~Zaremba, and P.~Abbeel, ``Overcoming
  {{Exploration}} in {{Reinforcement Learning}} with {{Demonstrations}},'' in
  \emph{2018 {{IEEE Int}}. {{Conf}}. {{Robot}}. {{Autom}}. {{ICRA}}}.\hskip 1em
  plus 0.5em minus 0.4em\relax {IEEE}, May 2018.

\bibitem{tai_socially_2018}
L.~Tai, J.~Zhang, M.~Liu, and W.~Burgard, ``Socially {{Compliant Navigation
  Through Raw Depth Inputs}} with {{Generative Adversarial Imitation
  Learning}},'' in \emph{2018 {{IEEE Int}}. {{Conf}}. {{Robot}}. {{Autom}}.
  {{ICRA}}}.\hskip 1em plus 0.5em minus 0.4em\relax {IEEE}, May 2018.

\bibitem{hoeller_learning_2021}
D.~Hoeller, L.~Wellhausen, F.~Farshidian, and M.~Hutter, ``Learning a {{State
  Representation}} and {{Navigation}} in {{Cluttered}} and {{Dynamic
  Environments}},'' \emph{IEEE Robot. Autom. Lett.}, vol.~6, no.~3, July 2021.

\bibitem{alahi_social_2016}
A.~Alahi, K.~Goel, V.~Ramanathan, A.~Robicquet, L.~{Fei-Fei}, and S.~Savarese,
  ``Social {{LSTM}}: {{Human Trajectory Prediction}} in {{Crowded Spaces}},''
  in \emph{2016 {{IEEE Conf}}. {{Comput}}. {{Vis}}. {{Pattern Recognit}}.
  {{CVPR}}}.\hskip 1em plus 0.5em minus 0.4em\relax {IEEE}, June 2016.

\bibitem{fernando_soft_2018}
T.~Fernando, S.~Denman, S.~Sridharan, and C.~Fookes, ``Soft + {{Hardwired}}
  attention: {{An LSTM}} framework for human trajectory prediction and abnormal
  event detection,'' \emph{Neural Networks}, vol. 108, Dec. 2018.

\bibitem{fujimoto_addressing_2018}
S.~Fujimoto, H.~Hoof, and D.~Meger, ``Addressing {{Function Approximation
  Error}} in {{Actor-Critic Methods}},'' in \emph{Proc. 35th {{Int}}. {{Conf}}.
  {{Mach}}. {{Learn}}.}\hskip 1em plus 0.5em minus 0.4em\relax {PMLR}, July
  2018.

\bibitem{shen_igibson_2021}
B.~Shen, F.~Xia, C.~Li, R.~{Mart{\'i}n-Mart{\'i}n}, L.~Fan, G.~Wang,
  C.~{P{\'e}rez-D'Arpino}, S.~Buch, S.~Srivastava, L.~Tchapmi, M.~Tchapmi,
  K.~Vainio, J.~Wong, L.~{Fei-Fei}, and S.~Savarese, ``{{iGibson}} 1.0: {{A
  Simulation Environment}} for {{Interactive Tasks}} in {{Large Realistic
  Scenes}},'' in \emph{2021 {{IEEERSJ Int}}. {{Conf}}. {{Intell}}. {{Robots
  Syst}}. {{IROS}}}, Sept. 2021.

\bibitem{coumans_pybullet_2016}
E.~Coumans and Y.~Bai, ``Pybullet: Physics simulation for games visual effects
  robotics and reinforcement learning,'' 2016.

\bibitem{alt_computing_1995}
H.~Alt and M.~Godau, ``Computing the fr\'echet distance between two polygonal
  curves,'' \emph{Int. J. Comput. Geom. Appl.}, vol.~05, no. 01n02, Mar. 1995.

\end{thebibliography}

\end{document}